# Learning with Scope,
## with Application to Information Extraction and Classification

**David M. Blei**
Computer Science Division
U.C. Berkeley
blei@cs.berkeley.edu

**J. Andrew Bagnell**
Robotics Institute
Carnegie Mellon University
dbagnell@ieee.org

**Andrew K. McCallum**
WhizBang Labs - Research
& Carnegie Mellon University
mccallum@whizbang.com

## Abstract

In probabilistic approaches to classification and information extraction, one typically builds a statistical model of words under the assumption that future data will exhibit the same regularities as the training data. In many data sets, however, there are scope-limited features whose predictive power is only applicable to a certain subset of the data. For example, in information extraction from web pages, word formatting may be indicative of extraction category in different ways on different web pages. The difficulty with using such features is capturing and exploiting the new regularities encountered in previously unseen data. In this paper, we propose a hierarchical probabilistic model that uses both local/scope-limited features, such as word formatting, and global features, such as word content. The local regularities are modeled as an unobserved random parameter which is drawn once for each local data set. This random parameter is estimated during the inference process and then used to perform classification with both the local and global features— a procedure which is akin to automatically retuning the classifier to the local regularities on each newly encountered web page. Exact inference is intractable and we present approximations via point estimates and variational methods. Empirical results on large collections of web data demonstrate that this method significantly improves performance from traditional models of global features alone.

## 1 Introduction

The web provides novel challenges to information extraction because structural regularities — its linkage, formatting, layout, and directories — can be powerful features that are necessarily ignored in traditional text learning. For example, to extract all the book titles on Amazon.com, one can rely on each title appearing in the same location and font on each book's page. Using this information in a machine learning context is difficult, however, because each web site has different structural regularities. Models trained on the structural information of one set of sites cannot be used for extraction or classification from another. In response, researchers have created tools for hand-tuning site-specific extractors (Cohen and Jensen, 2001).

The central problem with modeling site-specific regularities without hand-intervention is that most statistical models assume data to be independent and identically distributed (iid). As in the above example, certain disjoint subsets of the data may share identifiable and useful regularities which invalidate the iid assumption. Other examples of subsets that may have local regularities include patients from a particular hospital, voice sounds from a particular speaker, and vibration data from a particular airplane.

This paper presents scoped learning, a probabilistic framework that combines traditional iid features, such as word content, with scope limited features, such as formatting. We introduce a graphical model within this framework that exhibits two levels of scope, local and global. The global parameters are estimated from all the training data and the scope limited parameters, those shared for certain disjoint subsets of data, are modeled as unobserved random variables. It is important to note that this concept is recursively applicable to multiple levels of scope.

The intuitive procedure for using local features is to use information from the global (iid) features to infer the rules that govern the local information for a par-ticular subset of data. Continuing the example above, given that we know a certain number of book titles (global information), we can identify that they are all at the same location and font on the pages of Amazon.com. We can then reasonably infer that a previously unknown group of words, in that same location and font, is probably a book title. In the scoped learning framework, inference on a previously unseen subset of data naturally leads to exactly this process: first, estimate the local parameters from the observed global and local features; then, estimate the variable to be predicted (e.g., category) based on both sets of features and the inferred local parameters.

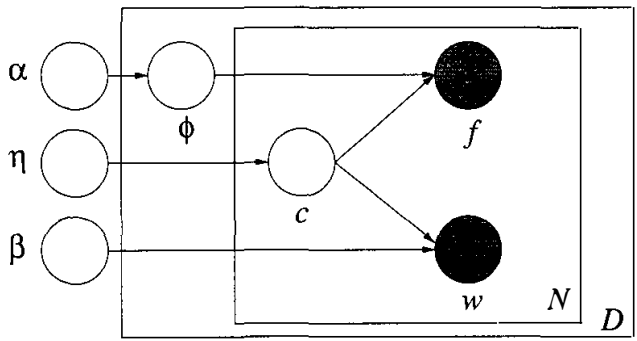

Figure 1: The graphical model representation of the scoped-learning model in a text domain. The square plates represent repetitions of documents ($D$) and the words in them ($N$). The (global) word content $w_n$ is generated independently of any parameters local to each document. The local formatting feature $f_n$ is generated dependent on the document-specific parameters $\phi$. These parameters, in turn, are generated once per document.

We describe generative and discriminative approaches to training and using the model for classification and extraction tasks. When data exhibits scope, we find significant gains in performance over traditional models which only use iid features.

## 2    A generative model with scope

In this section, we present a generative model of data which exhibits two kinds of scope: *global features* are governed by parameters which are shared by all data; *local features* are governed by parameters which are shared by particular subsets of the data.

A subset of data for which the local features exhibit the same regularity is called a *locale*. We assume that all future data will naturally be organized into identifiable locales. Note that the local regularities in the future data may be different from the local regularities in the data used for parameter estimation.

For simplicity of explanation, we will describe our model with reference to information extraction (IE) on web pages since the data naturally exhibits the scoped regularities which our model can exploit.

We cast IE as the classification problem of labeling certain individual words on a web page with different extraction categories such as "Job Title" or "Book Review". The data, comprising of word content and its corresponding formatting, can be easily divided into locales according to page boundaries. The assumption stated above holds since future data will naturally be organized into pages.

The individual terms on the page, without reference to their formatting, are the global features of the IE web data. This feature is a good predictor of extraction category in the same way for all documents. For example, the word "engineer" is probably part of a job title whether it appears on a page in www.amazon.com or www.google.com.

The local features are the formatting of the individual terms. These features, such as font size or color, can be a good predictor of category on a particular web page. The regularity which they capture, however, does not necessarily hold on other pages. For example, all the job titles on www.amazon.com may be green while all the job titles on www.google.com may be red.

### 2.1    Generative process and parameterization

To build a generative model, the data should be independent and identically distributed (iid). In the IE data, however, the word content (global features) and formatting (local features) are not iid since their joint distribution depends on the particular document to which each pair belongs. We can, however, model *documents* as iid by introducing an unobserved random parameter, drawn once for each document, that governs the local features. Positing this latent structure allows us to properly model the scope-limited features.

Denote the number of extraction categories by $K$, the size of the vocabulary by $V$, and number of possible values of the formatting feature (i.e., the formatting vocabulary) by $F$. Let an $N$ word document contain corresponding word content $\mathbf{w} = \{w_1, \ldots, w_N\}$, formatting $\mathbf{f} = \{f_1, \ldots, f_N\}$, and class labels $\mathbf{c} = \{c_1, \ldots, c_N\}$. We assume that a document is generated by the following process:

1. For each of the $K$ extraction categories:

   (a) Generate the formatting feature parameters $\phi_i$ from $p(\phi_i)$.

2. For each of the $N$ words in the document:

(a) Generate the $n$th class label $c_n$ from $p(c_n)$.

(b) Generate the $n$th word from $p(w_n|c_n)$.

(c) Generate the $n$th formatting feature from $p(f_n|c_n, \phi)$.

This models the following joint distribution over local parameters, class labels, words, and formatting features:

$$p(\phi, \mathbf{c}, \mathbf{w}, \mathbf{f}) = p(\phi) \prod_{n=1}^{N} p(c_n)p(w_n|c_n)p(f_n|c_n, \phi),$$

which is exhibited as a graphical model in Figure 1. The figure further illustrates the model parameters, $\alpha$, $\beta$, and $\eta$, which are suppressed in the equations. All probabilities are assumed to be conditioned on a point estimate of these parameters.

For each document and extraction category, $\phi_i$ is a random parameter to a multinomial over values of the formatting feature, i.e., $p(f_n|c_n, \phi) = \phi_{c_n f_n}$. Thus, each random parameter is a point on the $F$-simplex which we can parameterize by a Dirichlet distribution (recall that the Dirichlet is the exponential family distribution on the simplex). Note that we generate $K$ such parameters for each document.

Observe that there are no parameters which explicitly reference how the formatting features are generated for a particular document — only parameters on the random document-specific parameters which govern those features. This is the key element to the model since the formatting features will be generated by different (randomly-generated) parameters for different documents.

Finally, we assume the local features and global features are independent of each other given the class label. This is often a reasonable assumption. In IE, for example, the formatting of a word depends more on its role in the document (i.e., the extraction category) than what that word is in particular.

## 2.2 Parameter estimation

We fit the model by maximum likelihood estimation from a set of documents with observed class labels. Given the label, the global (word) features are independent of the local (formatting) features. From this independence, the parameters $p(w|c)$ and $p(c)$ can be estimated without using the formatting or page boundaries. Thus, we can train the class-conditional word distributions and class prior as a traditional generative probabilistic classifier such as naive Bayes.

In estimating the Dirichlet parameters on $p(\phi_i)$, we can quantify general trends in the local contexts (e.g., "boldness is usually indicative of class label" or "font size 12 is usually not indicative of class label"). We fix the Dirichlet parameters, however, to give a uniform distribution on the random local distributions. In doing so, we do not need to estimate any parameters with the formatting features of the training data. The formatting of future data, by the definition of the model, will be governed by different instances of $\phi$ than those that generated the training data.

## 2.3 Inference

Given an unlabeled web page, we would like to classify each word-formatting pair using the scoped-learning model described above. Inference on the class labels naturally leads to an intuitive method that uses both the word content and how it is formatted relative to the formatting of the rest of the document. First, use the uncertain labels given by the global information to infer the document-specific local parameters; then, refine the original (globally-estimated) labels using the local features.

Observe that, in the scoped-learning model, we need to simultaneously infer *all* the extraction labels of a document since they are dependent on each other through the unobserved local parameter. In particular, we compute the posterior distribution on the set of information labels $\mathbf{c}$ for some document $(\mathbf{w}, \mathbf{f})$. This can be computed, in principle, by marginalizing out the document-specific parameter $\phi$ and invoking Bayes rule:

$$p(\mathbf{c}|\mathbf{w}, \mathbf{f}) =$$
$$\frac{\int \prod_{n=1}^{N} p(w_n|c_n)p(f_n|c_n, \phi)p(c_n)p(\phi)d\phi}{\int \prod_{n=1}^{N} \sum_c p(w_n|c)p(f_n|c, \phi)p(c)p(\phi)d\phi}. \quad (1)$$

Exact computation of the integral is computationally infeasible because of the integral over the simplex of a product of sums in the denominator. We present two approaches to approximate inference, both of which operate on the model induced by the single document in question. The first approach is maximum likelihood estimation of $\phi$ on that model. The second, more Bayesian approach, uses variational methods to approximate the integral in Equation 1.

### 2.3.1 MAP estimates of the local parameter

By approximating $\phi$ with a point estimate $\hat{\phi}$, the posterior on $\mathbf{c}$ is tractable:

$$p(\mathbf{c}|\mathbf{w}, \mathbf{f}, \hat{\phi}) = \prod_{n=1}^{N} \frac{p(w_n|c_n)p(f_n|c_n, \hat{\phi})}{\sum_c p(w_n|c)p(f_n|c, \hat{\phi})}. \quad (2)$$

Furthermore, the labels for each word-formatting pair are conditionally independent given $\hat{\phi}$ and we can label

each pair with:

$$\hat{c}_n = \arg\max_c p(w_n|c) p(f_n|c, \hat{\phi}) p(c).$$

A natural point estimate for the local parameter is the posterior mode $\hat{\phi} = \arg\max_\phi p(\phi|\mathbf{f}, \mathbf{w})$. On the document in question, this corresponds to holding the global parameters fixed and finding a maximum likelihood estimate, using the observed formatting features, of the local parameters.

Since $\mathbf{c}$ is a set of unobserved random variables, we can use the Expectation-Maximization (EM) algorithm to maximize the expected log likelihood of the formatting features of the document:

$$\mathcal{L}_d(\phi) = \sum_{n=1}^{N} \sum_{c=1}^{K} p(c|w_n, f_n; \phi) \log p(f_n|c; \phi). \quad (3)$$

The E-Step computes the posterior distribution over extraction categories:

$$p^{(t+1)}(c|w_n, f_n; \phi) \propto p^{(t)}(f_n|c; \phi) p(w_n|c) p(c).$$

The M-Step finds the new estimate of $\hat{\phi} = p(f|c)$:

$$p^{(t+1)}(f|c; \phi) \propto \sum_{\{f_n : f_n = f\}} p^{(t)}(c|w_n, f_n),$$

where the notation under the sum indicates the set of formatting features $f_n$ which are equal to the value $f$. Alternating between these two steps guarantees that we take positive steps in Equation 3. Once converged, we use the resulting $\hat{\phi}$ to find the best value of $c_n$ for each word-formatting pair.

### 2.3.2 Variational approximation

An alternative approach is to use variational methods (Jordan et al., 1999) to approximate the true posterior distribution on the local parameters. A variational distribution can be used to provide a better approximation to the integrals in Equation 1 than the point estimate of the previous section.

For each new document, define a factorized variational distribution on the unobserved random variables $\mathbf{c}$ and $\phi$:

$$q(\mathbf{c}, \phi) = \prod_{k=1}^{K} q(\phi_k; \gamma_k) \prod_{n=1}^{N} q(c_n; \mu_n),$$

where $\gamma_k$ are a set of variational Dirichlet parameters for each extraction category and $\mu_n$ are a set of variational multinomial parameters over categories for each word-formatting pair. We optimize these parameters

to find the distribution that is as close as possible in KL divergence to the true posterior:

$$(\hat{\gamma}, \hat{\mu}) = \arg\min_{(\gamma, \mu)} \mathrm{KL}(q(\mathbf{c}, \phi; \gamma, \mu) \| p(\mathbf{c}, \phi|\mathbf{w}, \mathbf{f})).$$

Let $\gamma_{ij}$ be the Dirichlet parameter for the $j$th value of $f$ and the $i$th information category. The variational Dirichlet parameters are maximized by:

$$\gamma_{ij} = 1 + \sum_{\{f_n : f_n = j\}} q(c_n = i; \mu_n).$$

This update is similar to computing a posterior Dirichlet distribution with a uniform prior and expected counts under the variational distribution.

The variational multinomial parameters are updated by:

$$q(c_n; \mu_n) \propto p(c_n) p(w_n|c_n) \exp(\mathrm{E}[\log p(f_n; \phi); \gamma_c]).$$

This equation is similar to the E-step in Section 2.3.1 with $p(f_n|\hat{\phi}, c_n)$ replaced by the exponential of the expected value of its log under the posterior (variational) Dirichlet. That expectation is easily computable:

$$\mathrm{E}[\log p(f_n; \phi); \gamma_c] = \Psi(\gamma_{c f_n}) - \Psi(\textstyle\sum_f \gamma_{cf}),$$

where $\Psi$ is the digamma function. Iterating between the updates of $\gamma$ and $\mu$ will converge on a set of variational parameters which are close to the true posterior. Notice that the variational approximation maintains the same computational complexity as EM but may mitigate the problem of overfitting inside a locale.

We can view this attempt to approximate the integral over $\phi$ from a Bayesian point of view. A point estimate of the local parameters, such as the MAP estimate of section 2.3.1, is justified in large data limits where $p(\phi|\mathbf{w}, \mathbf{f})$ is peaked at a particular value of $\phi$. This large data assumption will generally be invalid because we specifically model small subsets of word-formatting pairs and thus, we can expect to gain from an attempt at proper integration. With this perspective, the update equations for the variational distribution follow the variational Bayes recipe given in Attias (2000).

## 3   A discriminative model with scope

In Section 2, we assume a generative model of the local (formatting) features. Often, such models place strong and unwarranted independence assumptions on the local features and it is desirable to learn a local discriminative model that captures $p(\mathbf{c}|\mathbf{f})$. In Figure 1, this corresponds to reversing the arc from local features $f$ to the category $c$ and changing the direction of the arrow on the random parameter $\phi$ to point to $c$.

As before, exact inference is intractable. We can find a point estimate for the optimal conditional parameters by maximizing the conditional log likelihood $p(\mathbf{w}|\mathbf{f}; \phi)$:

$$\log p(\mathbf{w}|\mathbf{f}; \phi) = \sum_{n=1}^{N} \log \sum_{c_n} p(c_n|f_n; \phi) p(w_n|c_n).$$

In the equation above, we implicitly assume a generative global classifier. To recover a completely discriminative solution, we use Bayes' rule to rewrite the conditional probability of the global features:

$$p(w_n|c_n) = p(c_n|w_n)p(w_n; \phi)/p(c_n; \phi).$$

This exposes a dependence of the likelihood on the marginal probability of the global features which, to be completely discriminative, we wish to avoid modeling. We can correct this by enforcing the constraint that $p(\mathbf{c})$ is constant with respect to $\phi$. Thus, $p(\mathbf{w})$ also is independent of $\phi$ and factorizes out of the likelihood. We then optimize the local parameters with respect to the following complete data likelihood:

$$\mathcal{L}_{\text{cond}}(\phi) = \sum_{n=1}^{N} \sum_{c_n} p(c_n|w_n, f_n) \log \frac{p(c_n|f_n; \phi)p(c_n|w_n)}{p(c_n)}.$$

The E-step, again, computes the posterior distribution:

$$p^{(t+1)}(c_n|w_n, f_n) \propto \frac{p^{(t)}(c_n|f_n)p(c_n|w_n)}{p(c_n)}.$$

The M-step corresponds to maximizing the weighted log likelihood:

$$J^{(t+1)} = \sum_{n=1}^{N} \sum_{c_n} p^{(t)}(c_n|w_n, f_n) \log p(c_n|f_n; \phi),$$

which is trivial to implement for most discriminative probabilistic classifiers (e.g., maximum entropy).

The conditional likelihood objective function $p(\mathbf{w}|\mathbf{f})$ has interesting information-theoretic content. The criteria is equivalent to maximizing an approximation to the mutual information:

$$\mathcal{I}(\mathbf{w}, \mathbf{f}; \phi) \approx \sum \hat{p}(w, f) \log \frac{p(w, f; \phi)}{p(w)p(f)},$$

where $\hat{p}(w, f)$ is the empirical distribution of the features. We can thus consider the class variables and the local parameters to be a "channel" where we try to recover as much information as possible about the global features from the local features.

## 4 Example and empirical results

### 4.1 Data

We test our scoped-learning model on two real-world data sets obtained from the web. The first is 1000 HTML documents that are automatically divided into sets of words with similar layout characteristics. Each group of words, called a text node, is hand-labeled as containing or not containing a job title. We split the data into two equal parts, a testing set and training set, keeping together the text nodes of a single document and documents of a single site. Given an unlabeled document from the testing set, we use a trained model to try to correctly label the text nodes. In this data, the word content are the global features and the word formatting are the local features (formatting features are easy to obtain in web data from the tags in the HTML source). A locale is a page since the job titles on each page tend to share formatting characteristics and all the data is naturally organized into pages.

Our second data set consists of 42,548 web pages from 330 web sites where each page is hand-labeled as being a press release or not being a press release. Given a new site, we use a trained model to try to identify the press-release pages within that site. The global features, again, are the words of the documents. The local features come from the URL because we expect that, within a particular web site, the URL structure is consistent with the function of the document to which it refers. For example, all press release pages of a particular company may be in a subdirectory company/news, or may be in files named similarly to pressrel20010402.html. The particular local features which we used were all the 4-grams of characters that appear in the URL. A locale in this dataset is a web site. Again, all the data is naturally organized into sites and we expect that press releases on each site will share the same URL regularities. As in the job title data set, we split the data equally into training and testing sets.

We note that the scoped-learning model which we use for both these data sets is slightly augmented from the one depicted in Figure 1. In particular, each class label is associated with multiple local and global features. Thus we have repetitions of the $w$ and $f$ nodes inside the repetition over class labels. The training and inference procedures, however, are almost exactly the same.

### 4.2 An example page

We describe an example web page from the job title corpus to aid in understanding the scoped-learning model presented here. Figure 2 shows a web page from the test set. Clearly, the job titles on this page have a consistent formatting.

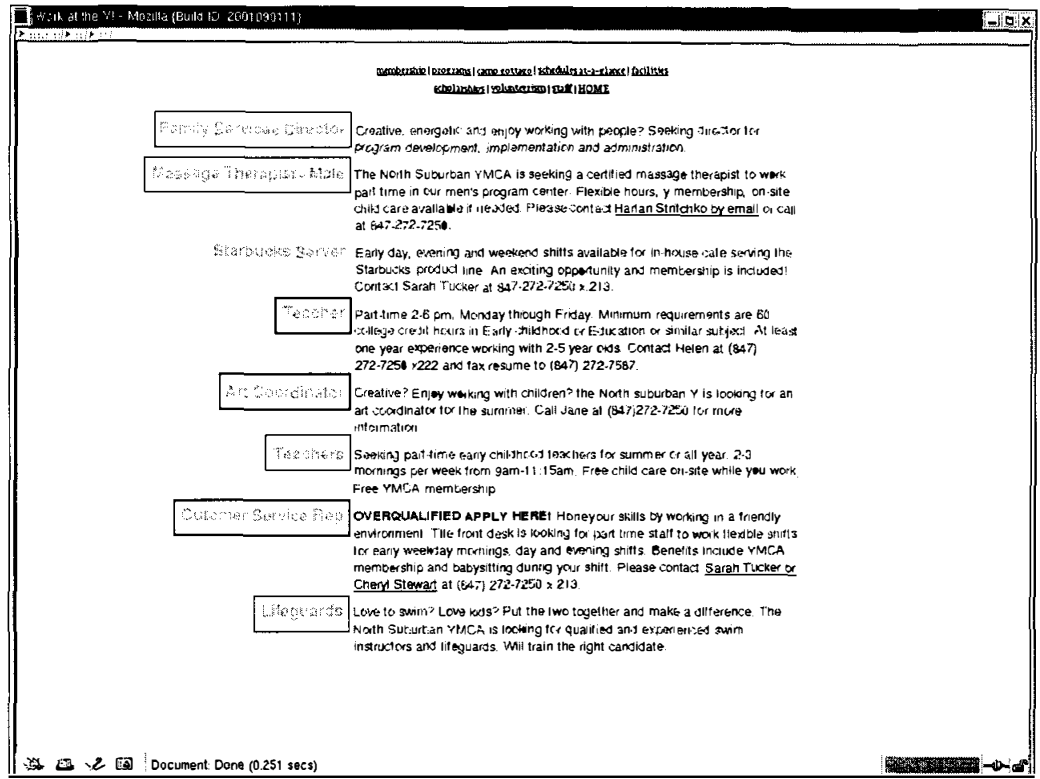

Figure 2: A sample web page from the job title dataset. The global classifier (depicted by the bold rectangles) correctly labeled 3 out of 8 jobs. Inference with our scoped-learning model (depicted by the thin and bold rectangles) correctly labeled 7 out of 8 jobs.

A simple naive Bayes classifier mislabels 5 of the 8 jobs on this page due to the lack of training data. In particular, it misses "Family Services Director", "Massage Therapist", "Starbucks Server", "Teachers", and "Lifeguard". For all these job titles, with the exception of "Starbucks Server", the classifier is relatively unsure of the label and only leans slightly away from classifying them correctly. Furthermore, the global classifier is very sure of the jobs which it correctly labels: "Teacher", "Art Coordinator", and "Customer Service Rep".

We applied the scoped-learning model to this document and learned, from the correctly labeled jobs, that the probability of boldness, the Arial font, and orange is very high when conditioned on the job title class label. Posterior inference on the class labels then yields the correct labeling for all but "Starbucks Server", giving a significant increase in classification accuracy. In the next section, we show that such improvement is typical on this data and these models can dramatically improve overall classification performance.

## 4.3 Quantitative results

Since both data sets correspond to binary classification problems, we quantify performance with precision and recall. Changing the threshold of probability with which to classify an example as being the positive class trades off classifying more negative examples as positive and finding more true positive examples (i.e., low precision, high recall) and missing more true positives but reducing the number of false positives (i.e., high precision, low recall).

The graph in Figure 3 (left) illustrates the performance of the naive Bayes classifier and the two generative variants of inference described in Section 2 on the job title data. The first variant is the MAP estimate of the local parameters learned via EM; it consistently dominates the performance of the global classifier. The second variant is the variational inference algorithm. This algorithm also dominates the global classifier and generally dominates the MAP estimate as well. At high levels of recall, which is an important part of the graph from a practical perspective, both algorithms that exploit the local features show significant improvement over the global classifier.

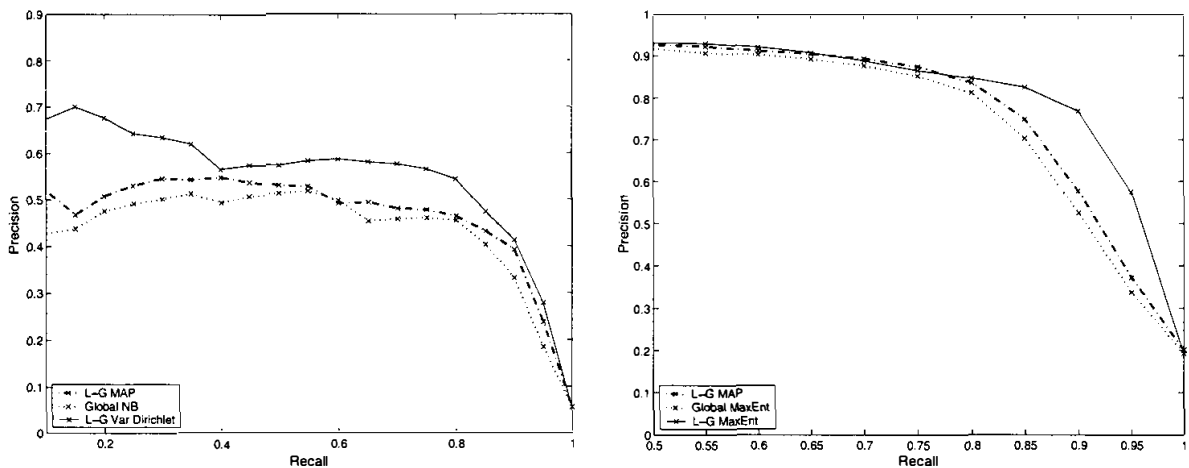

Figure 3: Scoped learning dominates the global classifier alone. (Left) A precision-recall curve that summarizes the the experimental results on the JOB-TITLE information extraction task. (Right) A precision-recall curve that summarizes the experimental results on the PRESS-RELEASE classification task.

The graph in Figure 3 (right) illustrates the performance of some of our algorithms on the press release data. In this case, the global classifier and scoped-learning discriminative classifiers are both maximum entropy classifiers with a Gaussian prior on the parameters. On this data, it is relatively easy to obtain high precision at low and moderate recall. The more interesting task, however, is to maintain such precision at higher recall. The scoped-learning models again outperform the global classifier at these levels of recall; they reduce error by more than one third at 90% recall and by more than half at 80% recall. Note that the local features are highly dependent and the discriminative approach outperforms the generative MAP estimate.

## 5   Related work

The central focus of this work is to emphasize the utility of features which exhibit scope-limited regularity and derive methods of learning this regularity in previously unseen data.

Blum and Mitchell's PAC-style co-training (1998) is one thread of related work. In this framework, two independent views of the data, each sufficient to predict the class label without error, can influence each other in making a final prediction. Although co-training does not exploit scope-limited features, its conditional independence assumption is similar to the one made here.

Taskar et al.'s work in probabilistic classification of relational data (2001) extends the notion of multiple views of a data set to multiple kinds of relation-ships and connections between the components of the data. The authors use probabilistic relational models (PRM's) to represent an entire data set as an interconnected graphical model. Inference on the unknown class variables implicitly exploits the regularities learned about different kinds of features which are present in the data.

In a sense, the model presented here can be cast as a PRM where each locale is represented as a separate group of nodes with a separate local parameter. This representation, and the maximum likelihood estimate of its parameters, exactly supports the MAP approximation in Section 2.3.1. By treating the local parameter as a random variable, however, we can explicitly treat each locale as iid and integrate out $\phi$. We showed in Section 4.3 that this leads to a significant increase in performance over the MAP estimate.

Our work is also related to learning supervised tasks using a combination of labeled and unlabeled data — particularly *transduction*, in which the unlabeled set is the test set (Vapnik, 1995). As in the transduction setting, we have no truly labeled data for a particular locale and need to learn the local parameters with uncertain labels obtained from an existing classifier. Previous work in this area, however, does not model locales, represent a difference between local and global features, or have the opportunity to use locales at training time to learn hyperparameters over local features.

Finally, Slattery and Mitchell build a system that automatically adjusts a trained classifier to the regularities discovered in the data which it is classifying (2000). This is similar to the scoped-learning model, but their method has no explicit notion of locale and is not designed for any features other than web page links. Furthermore, since it is not a probabilistic model, their method does not afford such benefits as integrating out parameter uncertainty.

# 6   Conclusion

We have presented a probabilistic model and training/inference procedures which can exploit both local and global features in various prediction tasks. On two large, real-world problems the model performs significantly better than the traditional global classifier. In comparing approximation methods, we find that the MAP estimate tends to overfit while integration via variational methods alleviates this issue. Furthermore, the discriminative approach can outperform the generative methods particularly when the local features are not independent.

Though we introduce the model with only two levels of feature scope, it is important to understand that the concept is recursively applicable to arbitrarily nested levels. This is necessary to capture the notion that some features may be meaningful, for example, on all English-language web sites, others within a corporate domain, and others within a single web page. Future work should leverage the more fine-grained distinction in feature scope which this framework admits.

Furthermore, we have developed and tested our models using simple mixtures as the underlying local and global models. It is possible to incorporate more sophisticated models of local and global features as well as structural relationships between class labels. For example, the models presented here treat the IE task as one of independent classification; a more natural model would be a finite-state sequence problem. Using scoped models in this way is straightforward by introducing a chain-structure to the class labels. With this change, the inference algorithms become a variant of standard inference routines on hidden Markov models.

An additional level of complexity can be gained by blurring the difference between local and global features. While we assume that the local and global features are disjoint, one can imagine models in which the same features appear in both the global and local settings and inference must determine which model has generated a particular occurrence of a feature. This would be useful, for example, in situations where word usage exhibits local regularities.

Finally, we have presented three approximate inference algorithms that are simple and efficient. Other methods, such as sampling or Expectation Propaga-tion (Minka, 2001), could also be used to estimate the posterior distribution on the local parameters; these approximations may lead to improved performance.

## Acknowledgments

This work was done while all three authors were working at WhizBang Labs during the summer 2001. We would like to thank Huan Chang for help with the job title data, William Cohen for advice and code related to generating local page features, Jonathan Baxter, Tom Minka, and Andrew Ng for insightful discussions, and Dallan Quass for inspirational work on extraction.